\documentclass[final]{cvpr}

\usepackage{times}
\usepackage{epsfig}
\usepackage{graphicx}
\usepackage{amsmath}
\usepackage{amssymb}

\usepackage{booktabs}
\usepackage{multirow}
\usepackage{subfigure}
\usepackage{threeparttable}
\usepackage{diagbox}
\usepackage{dsfont}
\usepackage{pifont}
\newcommand{\cmark}{\ding{51}}%
\newcommand{\xmark}{\ding{55}}%
\DeclareMathOperator*{\argmax}{arg\,max}
\DeclareMathOperator*{\argmin}{arg\,min}

\usepackage[pagebackref=true,breaklinks=true,colorlinks,bookmarks=false]{hyperref}

\def\cvprPaperID{5013} 
\def\confYear{CVPR 2021}

\begin{document}

\title{End-to-End Object Detection with Fully Convolutional Network}

\author{Jianfeng Wang\textsuperscript{\rm 1}\thanks{Equal contribution.}
\quad
Lin Song\textsuperscript{\rm 2}\footnotemark[1] \thanks{This work was done at Megvii Technology.}
\quad
Zeming Li\textsuperscript{\rm 1}
\quad
Hongbin Sun\textsuperscript{\rm 2}
\quad
Jian Sun\textsuperscript{\rm 1}
\quad
Nanning Zheng\textsuperscript{\rm 2}\\
\textsuperscript{\rm 1}Megvii Technology
\quad
\textsuperscript{\rm 2}Xi'an Jiaotong University\\
{\tt\small wangjianfeng@megvii.com
\quad
stevengrove@stu.xjtu.edu.cn}\\
{\tt\small \{lizeming,sunjian\}@megvii.com
\quad
\{hsun,nnzheng\}@mail.xjtu.edu.cn
}
}

\maketitle

\begin{abstract}
Mainstream object detectors based on the fully convolutional network has achieved impressive performance. While most of them still need a hand-designed non-maximum suppression (NMS) post-processing, which impedes fully end-to-end training. In this paper, we give the analysis of discarding NMS, where the results reveal that a proper label assignment plays a crucial role. To this end, for fully convolutional detectors, we introduce a Prediction-aware One-To-One (POTO) label assignment for classification to enable end-to-end detection, which obtains comparable performance with NMS. Besides, a simple 3D Max Filtering (3DMF) is proposed to utilize the multi-scale features and improve the discriminability of convolutions in the local region. With these techniques, our end-to-end framework achieves competitive performance against many state-of-the-art detectors with NMS on COCO and CrowdHuman datasets. The code is available at \url{https://github.com/Megvii-BaseDetection/DeFCN}.
\end{abstract}

\section{Introduction}

Object detection is a fundamental topic in computer vision, which predicts a set of bounding boxes with pre-defined category labels for each image.
Most of mainstream detectors~\cite{girshick2015fast, lin2017focal, redmon2016you, zhao2017pyramid} utilize some hand-crafted designs such as anchor-based label assignment and non-maximum suppression (NMS).
Recently, a quite number of methods~\cite{tian2019fcos, zhu2019feature, duan2019centernet} have been proposed to eliminate the pre-defined set of anchor boxes by using distance-aware and distribution-based label assignments.
Although they achieve remarkable progress and superior performance, there is still a challenge of discarding the NMS post-processing, which hinders the fully end-to-end training.

To tackle this issue, Learnable NMS~\cite{hosang2017learning}, Soft NMS~\cite{bodla2017soft} and other NMS variants~\cite{he2019bounding, liu2019adaptive, huang2020nms}, and CenterNet~\cite{duan2019centernet} are proposed to improve the duplicate removal, but they still do not provide an effective end-to-end training strategy.
Meanwhile, many approaches~\cite{stewart2016end, romera2016recurrent, park2015learning, ren2017end, salvador2017recurrent} based on recurrent neural networks have been introduced to predict the bounding box for each instance by using an autoregressive decoder.
These approaches give naturally sequential modeling for the prediction of bounding boxes.
But they are only evaluated on some small datasets without modern detectors, and the iterative manner makes the inference process inefficient.

\begin{figure}[t]
    \centering
    \includegraphics[width=\columnwidth]{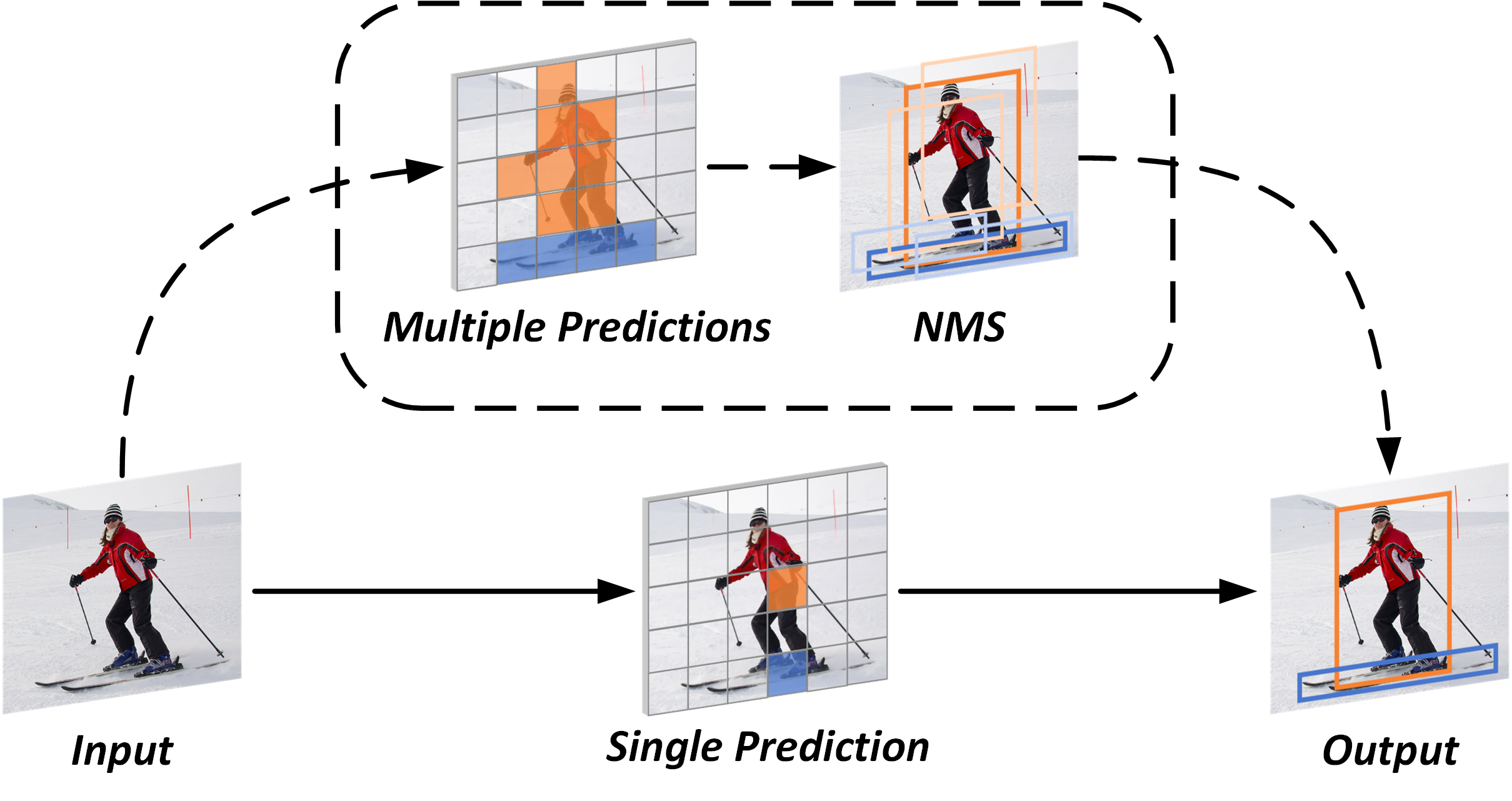}
    \caption{As shown in the dashed box, most detectors based on the fully convolutional network adopt multiple predictions and NMS post-processing for each instance. With the proposed prediction-aware one-to-one label assignment and 3D Max Filtering, our end-to-end detector can directly perform a single prediction for each instance without post-processing.}
    \label{fig:overview}
\end{figure}

Recently, DETR~\cite{carion2020end} introduces a bipartite matching based training strategy and transformers with the parallel decoder to enable end-to-end detection.
It achieves competitive performance against many state-of-the-art detectors.
However, DETR currently suffers from much longer training duration to coverage and relatively lower performance on the small objects.
To this end, this paper explores a new perspective: {\em could a fully convolutional network achieve competitive end-to-end object detection?}

In this paper, we attempt to answer this question in two dimensions, {\em i.e.}, label assignment and network architecture.
As shown in Fig.~\ref{fig:overview}, most of fully convolutional detectors~\cite{lin2017focal, tian2019fcos, zhang2020bridging, lin2017feature} adopt a one-to-many label assignment rule, {\em i.e.}, assigning many predictions as foreground samples for one ground-truth instance.
This rule provides adequate foreground samples to obtain a strong and robust feature representation.
Nevertheless, the massive foreground samples lead to duplicate predicted boxes for a single instance, which prevents end-to-end detection.
To demonstrate it, we first give an empirical comparison of different existing hand-designed label assignments.
We find that the one-to-one label assignment plays a crucial role in eliminating the post-processing of duplicate removal.
However, there is still a drawback in the hand-designed one-to-one assignment.
The fixed assignment could cause ambiguity issues and reduce the discriminability of features, since the predefined regions of an instance may not be the best choice~\cite{kim2020probabilistic} for training.
To solve this issue, we propose a prediction-aware one-to-one (POTO) label assignment, which dynamically assigns the foreground samples according to the quality of classification and regression simultaneously.

Furthermore, for the modern FPN based detector~\cite{tian2019fcos}, the extensive experiment demonstrates that the duplicate bounding boxes majorly come from the nearby regions of the most confident prediction across adjacent scales.
Therefore, we design a 3D Max Filtering (3DMF), which can be embedded into the FPN head as a differentiable module.
This module could improve the discriminability of convolution in the local regions by using a simple 3D max filtering operator across adjacent scales.
Besides, to provide adequate supervision for feature representation learning, we modify a one-to-many assignment as an auxiliary loss.

With the proposed techniques, our end-to-end detection framework achieves competitive performance against many state-of-the-art detectors.
On COCO~\cite{lin2014microsoft} dataset, our end-to-end detector based on FCOS framework~\cite{tian2019fcos} and ResNeXt-101~\cite{xie2017aggregated} backbone remarkably outperforms the baseline with NMS by {\bf 1.1\%} mAP.
Furthermore, our end-to-end detector is more robust and flexible for crowded detection.
To demonstrate the superiority in the crowded scenes, we construct more experiments on CrowdHuman~\cite{shao2018crowdhuman} dataset.
Under the ResNet-50 backbone, our end-to-end detector achieves {\bf 3.0\%} AP$_{50}$ and {\bf 6.0\%} mMR absolute gains over FCOS baseline with NMS.

\section{Related Work}

\subsection{Fully Convolutional Object Detector}

Owing to the success of convolution networks~\cite{he2016deep, song2019learnable, song2019tacnet, song2020rethinking, li2020learning, zhang2019glnet, zhang2019latentgnn}, object detection has achieved tremendous progress during the last decade.
Modern one-stage~\cite{lin2017focal, liu2016ssd, redmon2018yolov3, song2020fine, qiu2020borderdet, ge2021lla} or two-stage detectors~\cite{ren2015faster, lin2017feature, cai2018cascade} heavily rely on the anchors or anchor-based proposals.
In these detectors, the anchor boxes are made up of pre-defined sliding windows, which are assigned as foreground or background samples with bounding box offsets.
Due to the hand-designed and data-independent anchor boxes, the training targets of anchor-based detectors are typically sub-optimal and require careful tuning of hyper-parameters.
Recently, FCOS~\cite{tian2019fcos} and CornerNet~\cite{law2018cornernet} give a different perspective for fully convolutional detectors by introducing an anchor-free framework.
Nevertheless, these frameworks still need a hand-designed post-processing step for duplicate removal, {\em i.e.}, non-maximum suppression (NMS).
Since NMS is a heuristic approach and adopts a constant threshold for all the instances, it needs carefully tuning and might not be robust, especially in crowded scenes.
In contrast, based on the anchor-free framework, this paper proposes a prediction-aware one-to-one assignment rule for classification to discard the non-trainable NMS.

\subsection{End-to-End Object Detection}

To achieve end-to-end detection, many approaches are explored in the previous literature.
Concretely, in the earlier researches, numerous detection frameworks based on recurrent neural networks~\cite{stewart2016end, romera2016recurrent, park2015learning, ren2017end, salvador2017recurrent} attempt to produce a set of bounding boxes directly.
Albeit they allow end-to-end learning in principle, they are only demonstrated effectiveness on some small datasets and not against the modern baselines~\cite{tian2019fcos, ghiasi2019fpn}.
Meanwhile, Learnable NMS~\cite{hosang2017learning} is proposed to learn duplicate removal by using a very deep and complex network, which achieves comparable performance against NMS.
But it is constructed by discrete components and does not give an effective solution to realize end-to-end training.
Recently, the relation network~\cite{hu2018relation} and DETR~\cite{carion2020end} apply the attention mechanism to object detection, which models pairwise relations between different predictions.
By using one-to-one assignment rules and direct set losses, they do not need any additional post-processing steps.
Nevertheless, when performing massive predictions, these methods require highly expensive cost, making them not appropriate for the dense prediction frameworks.
Due to the lack of image prior and multi-scale fusion mechanism, DETR also suffers from much longer training duration than mainstream detectors and lower performance on the small objects.
Different from the approaches mentioned above, our method is the first to enable end-to-end object detection based on a fully convolutional network.

\begin{table*}[htbp]
    \centering
    \caption{The comparison of different label assignment rules for end-to-end object detection on COCO {\em val} set. $\Delta$ indicates the gap between with and without NMS. `Aux' is the proposed auxiliary loss. All models are based on ResNet-50 backbone with 180k training iterations.}
    \resizebox{\linewidth}{!}{\begin{threeparttable}
    \begin{tabular}{l|l|l|ccc|ccc}
        \toprule
        \multirow{2}{*}{Assignment} &
        \multirow{2}{*}{Rule} &
        \multirow{2}{*}{Method} & \multicolumn{3}{c|}{mAP} & \multicolumn{3}{c}{mAR} \\
         &  &  & w/ NMS & w/o NMS & $\Delta$ & w/ NMS & w/o NMS & $\Delta$ \\
        \midrule
        One-to-many & Hand-designed & FCOS~\cite{tian2019fcos} baseline~\tnote{*} & 40.5 & 12.1 & -28.4 & 58.3 & 52.8 & -5.5 \\
        \midrule
        \multirow{4}{*}{One-to-one} & \multirow{2}{*}{Hand-designed} & Anchor & 37.2 & 35.8 & -1.4 & 57.0 & 59.2 & +2.2 \\
         &  & Center & 37.2 & 33.6 & -3.6 & 57.8 & 59.7 & +1.9 \\ \cmidrule{2-9}
         & \multirow{3}{*}{Prediction-aware} & Foreground loss & 38.3 & 37.1 & -1.2 & 58.6 & \bf 61.4 & \bf +2.8 \\
         &  & POTO & 38.6 & 38.0 & -0.6 & 57.9 & 60.5 & +2.6 \\
         &  & POTO+3DMF & 40.0 & 39.8 & -0.2 & 58.8 & 60.9 & +2.1 \\
        \midrule
        Mixture~\tnote{**} & Prediction-aware & POTO+3DMF+Aux & \bf 41.2 & \bf 41.1 & \bf -0.1 & \bf 58.9 & 61.2 & +2.3 \\
        \bottomrule
    \end{tabular}
    \begin{tablenotes}
        \small
        \item[*] We remove its centerness branch to achieve a head-to-head comparison.
        \item[**] We adopt a one-to-one assignment in POTO and a one-to-many assignment in the auxiliary loss, respectively.
    \end{tablenotes}
    \label{tab:one-to-one}
    \end{threeparttable}}
\end{table*}

\section{Methodology}

\subsection{Analysis on Label Assignment}

To reveal the effect of label assignment on end-to-end object detection, we construct several ablation studies of conventional label assignments on COCO~\cite{lin2014microsoft} dataset.
As shown in Tab.~\ref{tab:one-to-one}, all the experiments are based on FCOS~\cite{tian2019fcos} framework, whose centerness branch is removed to achieve a head-to-head comparison.
The results demonstrate the superiority of one-to-many assignment on feature representation and the potential of one-to-one assignment on discarding the NMS.
The detailed analysis is elaborated in the following sections.

\subsubsection{One-to-many Label Assignment}

Since the NMS post-processing is widely adopted in dense prediction frameworks~\cite{lin2017feature, lin2017focal, zhu2019feature, zhang2019freeanchor, tian2019fcos, zhang2020bridging, zhu2020autoassign, qiu2020borderdet, ge2021lla}, one-to-many label assignment becomes a conventional way to assign training targets.
The adequate foreground samples lead to a strong and robust feature representation.
However, when discarding the NMS, due to the redundant foreground samples of one-to-many label assignment, the duplicate false-positive predictions could cause a dramatic drop in performance, {\em e.g.}, 28.4\% mAP absolute drop on FCOS~\cite{lin2017focal} baseline.
In addition, the reported mAR in Tab.~\ref{tab:one-to-one} indicates the recall rates for the predictions of the top 100 scores. Without NMS, the one-to-many assignment rule leads to numerous duplicate predictions with high scores, thus reducing the recall rate.
Therefore, the detector is hard to achieve competitive end-to-end detection by relying only on the one-to-many assignment.

\subsubsection{Hand-designed One-to-one Label Assignment}
\label{sec:one-to-one_assign}

MultiBox~\cite{szegedy2014scalable} and YOLO~\cite{redmon2016you} demonstrate the potential in applying the one-to-one label assignment to a dense prediction framework.
In this paper, we evaluate two one-to-one label assignment rules to reveal the undergoing connection with discarding NMS.
These rules are modified by two widely-used one-to-many label assignments: {\em Anchor} rule and {\em Center} rule.
Concretely, {\em Anchor} rule is based on RetinaNet~\cite{lin2017focal}, each ground-truth instance is only assigned to the anchor with the maximum Intersection-over-Union (IoU). 
{\em Center} rule is based on FCOS~\cite{tian2019fcos}, each ground-truth instance is only assigned to the pixel closest to the center of the instance in the pre-defined feature layer.
Besides, other anchors or pixels are set as background samples.


As shown in Tab.~\ref{tab:one-to-one}, compared with the one-to-many label assignment, the one-to-one label assignment allows the fully convolutional detectors without NMS to greatly reduce the gap between with and without NMS and achieve reasonable performance.
For instance, the detector based on {\em Center} rule achieves 21.5\% mAP absolute gains over the FCOS baseline.
Besides, as it avoids the error suppression of the NMS in complex scenes, the recall rate is further increased.
Nevertheless, there still exist two unresolved issues.
First, when one-to-one label assignment is applied, the performance gap between detectors with and without NMS remains non-negligible.
Second, due to the less supervision for each instance, the performance of the one-to-one label assignment is still inferior to the FCOS baseline.

\begin{figure*}[t]
    \centering
    \includegraphics[width=0.9\textwidth]{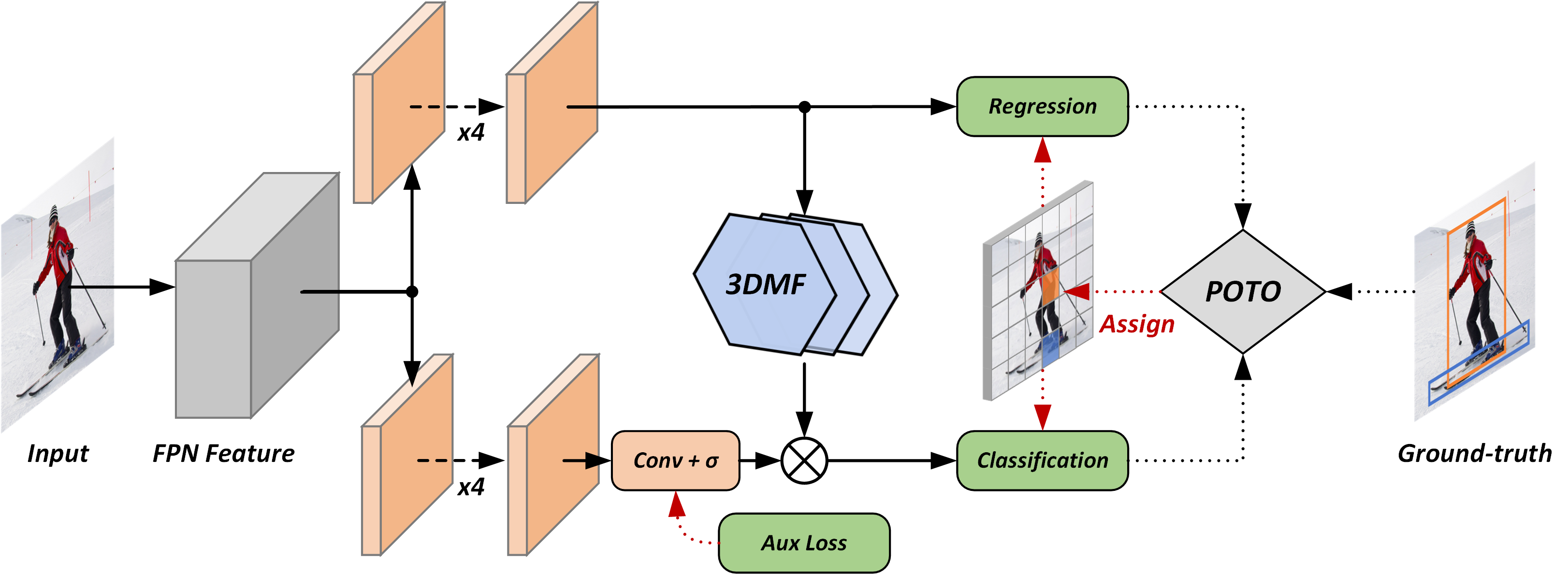}
    \caption{The diagram of the head with 3D Max Filtering (3DMF) in a FPN stage. `POTO' indicates the proposed Prediction-aware One-to-one Label Assignment rule to achieve end-to-end detection. 
    `Conv + $\sigma$' denotes a convolution layer followed by a sigmoid function~\cite{han1995influence}, which outputs coarsely classification scores.
    `Aux Loss' is the proposed auxiliary loss to improve feature representation. The dotted lines are used to highlight the additional components in the training phase, which are abandoned in the inference phase.}
    \label{fig:head}
\end{figure*}

\subsection{Our Methods}

In this paper, to enable competitive end-to-end object detection, we propose a mixture label assignment and a new 3D Max Filtering (3DMF).
The mixture label assignment is made up of the proposed prediction-aware one-to-one  (POTO) label assignment and a modified one-to-many label assignment (auxiliary loss).
With these techniques, our end-to-end framework can discard the NMS post-processing and keep the strong feature representation.

\subsubsection{Prediction-aware One-to-one Label Assignment}

The hand-designed one-to-one label assignment follows a fixed rule.
However, this rule may be sub-optimal for various instances in complex scenes, {\em e.g.}, {\em Center} rule for an eccentric object~\cite{kim2020probabilistic}.
Thus if the assignment procedure is forced to assign the sub-optimal prediction as the unique foreground sample, the difficulty for the network to converge could be dramatically increased, leading to more false-positive predictions.
To this end, we propose a new rule named Prediction-aware One-To-One (POTO) label assignment by dynamically assigning samples according to the quality of predictions.

Let $\Psi$ denotes the index set of all the predictions.
$G$ and $N$ correspond to the number of ground-truth instances and predictions, respectively, where typically $G \ll N$ in dense prediction detectors.
$\hat{\pi} \in \Pi_G^N$ indicates a $G$-permutation of $N$ predictions.
Our POTO aims to generate a suitable permutation $\hat{\pi}$ of predictions as the foreground samples.
The training loss is formulated as Eq.~\ref{eq:loss}, which consists of the foreground loss $\mathcal{L}_{\mathit{fg}}$ and the background loss $\mathcal{L}_{\mathit{bg}}$.
\begin{gather}
    \mathcal{L} = \sum_i^G \mathcal{L}_\mathit{fg} \left( \hat{p}_{\hat{\pi}(i)}, \hat{b}_{\hat{\pi}(i)} \mid c_i, b_i \right) + \sum_{j \in \Psi \setminus \mathcal{R}(\hat{\pi})} \mathcal{L}_\mathit{bg} \bigl( \hat{p}_j \bigr),
    \label{eq:loss}
\end{gather}
where $\mathcal{R}(\hat{\pi})$ denotes the corresponding index set of the assigned foreground samples.
For the $i$-th ground-truth, $c_i$ and $b_i$ are its category label and bounding box coordinates, respectively.
While for the $\hat{\pi}(i)$-th prediction, $\hat{p}_{\hat{\pi}(i)}$ and $\hat{b}_{\hat{\pi}(i)}$ correspond to its predicted classification scores and predicted box coordinates, respectively.

To achieve competitive end-to-end detection, we need to find a suitable label assignment $\hat{\pi}$.
As shown in Eq.~\ref{eq:argmin_loss}, previous works~\cite{erhan2014scalable, carion2020end} treat it as a bipartite matching problem by using foreground loss~\cite{lin2017focal, rezatofighi2019generalized} as the matching cost, which can be rapidly solved by the Hungarian algorithm~\cite{stewart2016end}.
\begin{gather}
    \hat{\pi} = \argmin_{\pi \in \Pi_G^N} \sum_i^G \mathcal{L}_\mathit{fg}\left( \hat{p}_{\hat{\pi}(i)}, \hat{b}_{\hat{\pi}(i)} \mid c_i, b_i \right). \label{eq:argmin_loss}
\end{gather}
However, foreground loss typically needs additional weights to alleviate optimization issues, {\em e.g.}, unbalanced training samples and joint training of multiple tasks.
As shown in Tab.~\ref{tab:one-to-one}, this property makes the training loss not the optimal choice for the matching cost.
Therefore, as presented in Eq.~\ref{eq:bipartite} and Eq.~\ref{eq:quality}, we propose a more clean and effective formulation (POTO) to find a better assignment.
\begin{gather}
    \hat{\pi} = \argmax_{\pi \in \Pi_G^N} \sum_i^G Q_{i,\pi(i)}, \label{eq:bipartite}\\
    \begin{aligned}
        \text{where}~Q_{i,\pi(i)} =& \underbrace{\mathds{1}\left[ \pi(i) \in \Omega_i \right]}_{\text{spatial prior}} ~\cdot~ \underbrace{{\Bigl( \hat{p}_{\pi(i)}(c_i) \Bigr)}^{1-\alpha}}_{\text{classification}} ~\cdot\\
        &\underbrace{{\Bigl( \mathrm{IoU} \bigl( b_i, \hat{b}_{\pi(i)} \bigr) \Bigr)}^\alpha}_{\text{regression}}.
    \label{eq:quality}
    \end{aligned}
\end{gather}
Here $Q_{i,\pi(i)} \in [0, 1]$ represents the proposed matching quality of the $i$-th ground-truth with the $\pi(i)$-th prediction.
It considers the spatial prior, the confidence of classification, and the quality of regression simultaneously.
$\Omega_i$ indicates the set of candidate predictions for $i$-th ground-truth, {\em i.e.}, spatial prior.
The spatial prior is widely used in the training phase~\cite{lin2017feature, lin2017focal, zhu2019feature, zhang2019freeanchor, tian2019fcos, zhang2020bridging}.
For instance, the center sampling strategy is adopted in FCOS~\cite{tian2019fcos}, which only considers the predictions in the central portion of the ground-truth instance as foreground samples.
We also apply it in POTO to achieve higher performance, but it is not necessary for discarding NMS (more details refer to Sec.~\ref{sec:label_assign}).
To achieve balance, we define the quality by the weighted geometric mean of classification score $\hat{p}_{\pi(i)}(c_i)$ and regression quality $\mathrm{IoU} \bigl( b_i, \hat{b}_{\pi(i)} \bigr)$ in Eq.~\ref{eq:quality}.
The hyper-parameter $\alpha \in [0, 1]$ adjusts the ratio between classification and regression, where $\alpha = 0.8$ is adopted by default and more ablation studies are elaborated in Sec.~\ref{sec:label_assign}.
As shown in Tab.~\ref{tab:one-to-one}, POTO not only narrows the gap with NMS but also improves the performance.

\begin{table}[t]
    \centering
    \caption{Comparison of different configurations for NMS post-processing on COCO {\em val} set. `Across scales' indicates applying NMS to the multiple adjacent stages of the feature pyramid network. `Spatial range' denotes the spatial range for duplicate removal in each scale.}
    \begin{tabular}{l|c|c|c}
        \toprule
        Model & Across scales & Spatial range & mAP \\
        \midrule
        \multirow{6}{*}{FCOS~\cite{tian2019fcos}} & \multirow{3}{*}{\xmark} & $1\times1$ & 19.0 \\  
         &  & $3\times3$ & 37.4 \\  
         &  & $5\times5$ & 39.2 \\  
        \cmidrule{2-4}
         & \xmark & \multirow{2}{*}{$\infty \times \infty$} & 39.2 \\  
         & \cmark &  & 40.9 \\  
        \bottomrule
    \end{tabular}
    \label{tab:scale-range-nms}
\end{table}

\subsubsection{3D Max Filtering}
\label{sec:3d_max}

In addition to the label assignment, we attempt to design an effective architecture to realize more competitive end-to-end detection.
To this end, we first reveal the distribution of duplicate predictions.
As shown in Tab.~\ref{tab:scale-range-nms}, for a modern FPN based detector~\cite{tian2019fcos}, the performance has a noticeable degradation when applying the NMS to each scale separately.
Moreover, we find that the duplicate predictions majorly come from the nearby spatial regions of the most confident prediction.
Therefore, we propose a new module called 3D Max Filtering (3DMF) to suppress duplicate predictions.

Convolution is a linear operation with translational equivariance, which produces similar outputs for similar patterns at different positions.
However, this property has a great obstacle to duplicate removal, since different predictions of the same instance typically have similar features~\cite{lin2017focal} for the dense prediction detectors.
Max filter is a rank-based non-linear filter~\cite{sonka2014image}, which could be used to compensate for the discriminant ability of convolutions in a local region.
Besides, max filter has also been utilized in the key-point based detectors, {\em e.g.}, CenterNet~\cite{zhou2019objects} and CornerNet~\cite{law2018cornernet}, as a new post-processing step to replace the non-maximum suppression.
It demonstrates some potentials to perform duplicate removal, but the non-trainable manner hinders the effectiveness and end-to-end training.
Meanwhile, the max filter only considers the single-scale feature, which is not appropriate for the widely-used FPN based detectors~\cite{lin2017focal, tian2019fcos, zhang2020bridging}.

\begin{figure}[t]
    \centering
    \includegraphics[width=0.75\columnwidth]{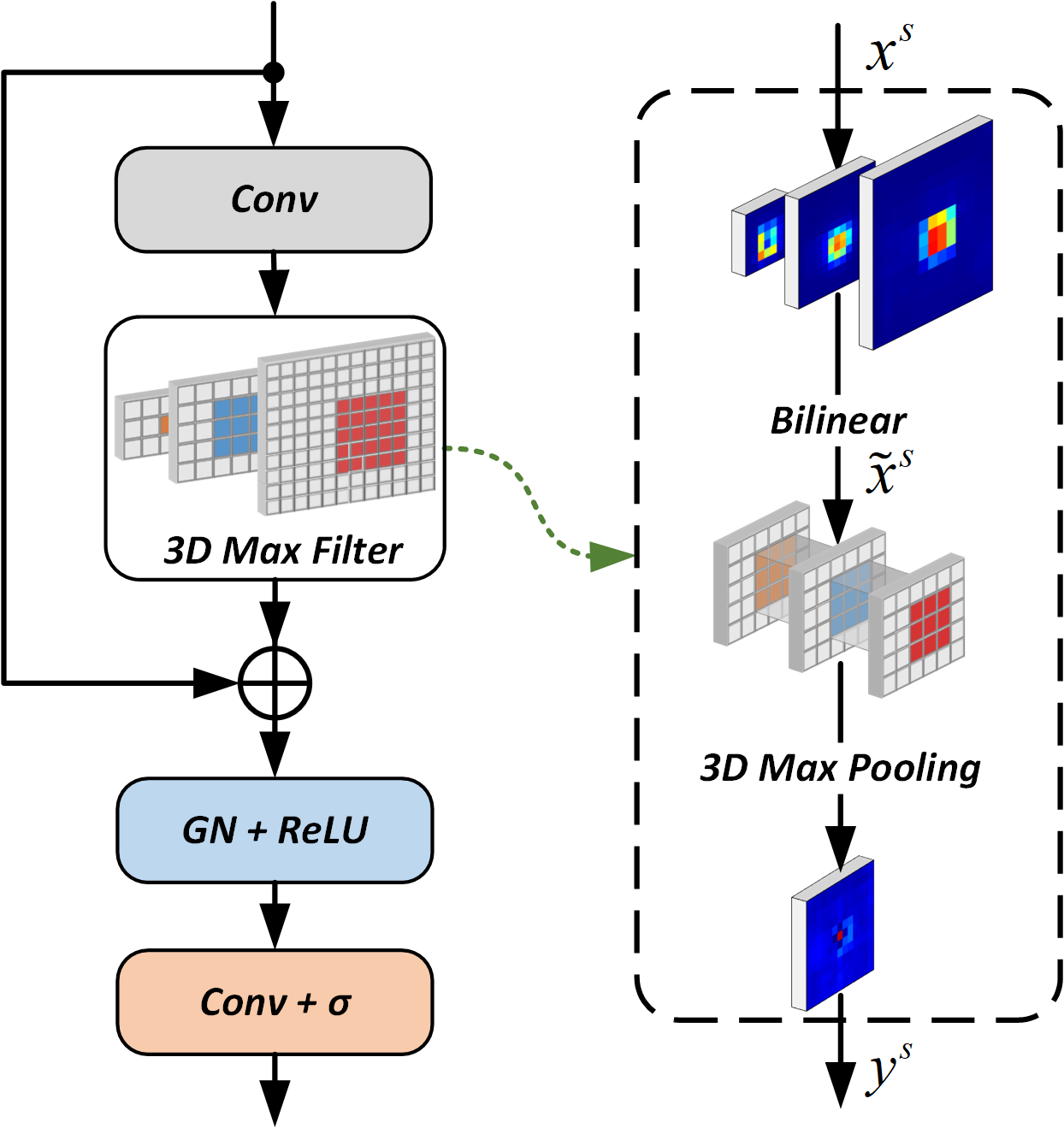}
    \caption{The diagram of 3D Max Filtering. The detailed procedure of 3D max filtering is illustrated in the dashed box. `GN' and `$\sigma$' indicate the group normalization~\cite{wu2018group} and the sigmoid activation function, respectively.}
    \label{fig:max3d}
\end{figure}

Therefore, we extend the max filter to a multi-scale version, called 3D Max Filtering, which transforms the features in each scale of FPN.
The 3D Max Filtering is respectively adopted in each channel of a feature map.
\begin{gather}
    \tilde{x}^s =
    \left\{ \tilde{x}^{s,k}:=\mathop{\mathrm{Bilinear}}\limits_{x^s}(x^k) \mid \forall k\in \left[s - \frac{\tau}{2}, s + \frac{\tau}{2}\right] \right\}.
\label{eq:max_interpolate}
\end{gather}
Specifically, as shown in Eq.~\ref{eq:max_interpolate}, given an input feature $x^s$ in the scale $s$ of FPN, we first adopt the bilinear operator~\cite{press2007numerical} to interpolate the features from $\tau$ adjacent scales as the same size of input feature $x^s$.
\begin{gather}
    y^s_i = \max_{k \in \left[s - \frac{\tau}{2}, s + \frac{\tau}{2}\right]}\max_{j \in \mathcal{N}^{\phi\times \phi}_i} \tilde{x}^{s,k}_j.
\label{eq:max_value}
\end{gather}
As shown in Eq.~\ref{eq:max_value}, for a spatial location $i$ in scale $s$, the maximum value $y^s_i$ is then obtained in a pre-defined 3D neighbour tube with $\tau$ scales and $\phi \times \phi$ spatial distance.
This operation can be easily implemented by a highly efficient 3D max-pooling operator~\cite{paszke2019pytorch}.

\begin{figure*}[ht]
\centering
\includegraphics[width=0.98\linewidth]{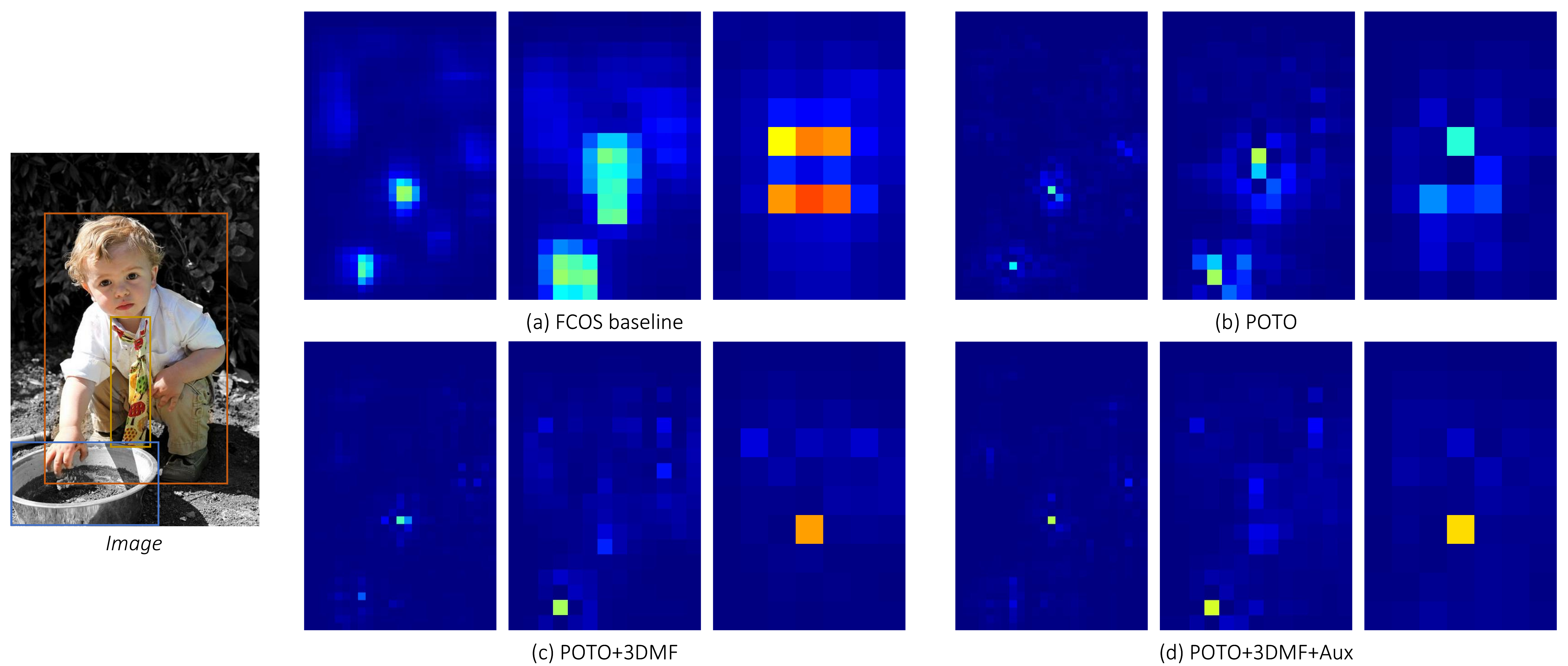}
\caption{
Visualization of the predicted classification scores from different approaches. The input image has three instances of different scales, {\em i.e.}, person, tie and pot. The heatmaps from left to right of each approach correspond to the score map in the FPN stage `P5', `P6' and `P7', respectively. `Aux' indicates the proposed auxiliary loss. Our POTO based detector significantly suppresses the duplicate predictions against the vanilla FCOS framework. The 3DMF enhances the distinctiveness of the local region across adjacent scales. Besides, the auxiliary loss can further improve the feature representation.}
\label{fig:coco_vis}
\end{figure*}

Furthermore, to embed the 3D Max Filtering into the existing frameworks and enable end-to-end training, we propose a new module, as shown in Fig.~\ref{fig:max3d}.
This module leverages the max filtering to select the predictions with the highest activation value in a local region and could enhance the distinction with other predictions, which is further verified in Sec.~\ref{sec:visualization}.
Owing to this property, as shown in Fig.~\ref{fig:head}, we adopt the 3DMF to refine the coarsely dense predictions and suppress the duplicate predictions.
Besides, all the modules are constructed by simple differentiable operators and only have slightly computational overhead.

\subsubsection{Auxiliary Loss}
\label{sec:aux_loss}

In addition, when using the NMS, as shown in Tab.~\ref{tab:one-to-one}, the performance of POTO and 3DMF is still inferior to the FCOS baseline.
This phenomenon may be attributed to the fact that one-to-one label assignment provides less supervision, making the network difficult to learn the strong and robust feature representation~\cite{szegedy2015going}.
It could further reduce the discrimination of classification, thus causing a decrease in performance.
To this end, motivated by many previous works~\cite{szegedy2015going, zhao2017pyramid, zhao2019object}, we introduce an auxiliary loss based on one-to-many label assignment to provide adequate supervision, which is illustrated in Fig.~\ref{fig:head}.

Similar to ATSS~\cite{zhang2020bridging}, our auxiliary loss adopts the focal loss~\cite{lin2017focal} with a modified one-to-many label assignment.
Specifically, the one-to-many label assignment first takes the top-9 predictions as candidates in each FPN stage, according to the proposed matching quality in Eq.~\ref{eq:quality}.
It then assigns the candidates as foreground samples whose matching qualities beyond a statistical threshold.
The statistical threshold is calculated by the summation of the mean and the standard deviation of all the candidate matching qualities.
In addition, different forms of one-to-many label assignment for the auxiliary loss are elaborately reported in the supplementary material.



\section{Experiments}


\subsection{Implement Detail}

As same as FCOS~\cite{tian2019fcos}, our detector adopts a pair of 4-convolution heads for classification and regression, respectively.
The output channel numbers of the first convolution and the second convolution in 3DMF are 256 and 1, respectively.
All the backbones are pre-trained on the ImageNet dataset~\cite{deng2009imagenet} with frozen batch normalizations~\cite{ioffe2015batch}.
In the training phase, input images are reshaped so that their shorter side is 800 pixels.
All the training hyper-parameters are identical to the 2x schedule (180k iterations) in the Detectron2~\cite{wu2019detectron2} if not specifically mentioned.

\subsection{Ablation Studies on COCO}

\subsubsection{Visualization}
\label{sec:visualization}

As shown in Fig.~\ref{fig:coco_vis}, we present the visualization of the classification scores from the FCOS baseline and our proposed framework.
For a single instance, the FCOS baseline with one-to-many assignment rule outputs massive duplicate predictions, which are highly activated and have comparable activating scores with the most confident one.
These duplicate predictions are evaluated as false-positive samples and greatly affect performance.
In contrast, by using the proposed POTO rule, the scores of duplicate samples are significantly suppressed.
This property is crucial for the detector to achieve direct bounding box prediction without NMS.
Moreover, with the proposed 3DMF module, this property is further enhanced, especially in the nearby regions of the most confident prediction.
Besides, since the 3DMF module introduces the multi-scale competitive mechanism, the detector can well perform unique predictions across different FPN stages, {\em e.g.}, an instance in the Fig.~\ref{fig:coco_vis} has single highly activated scores in various stages.


\subsubsection{Prediction-Aware One-to-One Label Assignment}
\label{sec:label_assign}


\noindent\textbf{Spatial prior.} 
As shown in Tab.~\ref{tab:assign}, for the spatial range of assignment, the center sampling strategy is relatively superior to the inside box and global strategies on the COCO dataset.
It reflects that the prior knowledge of images is essential in the real world scenario.

\begin{table}[t]
    \centering
    \caption{Results of POTO with different configurations of $\alpha$ and spatial prior on COCO {\em val} set. $\alpha = 0$ is equivalent to considering classification alone, $\alpha = 1$ is equivalent to considering regression alone. `center sampling' and `inside box' both follow FCOS~\cite{tian2019fcos}. `/' is used to distinguish between results without and with NMS.}
    \begin{tabular}{c|ccc}
        \toprule
        $\alpha$ & center sampling & inside box & global \\
        \midrule
        0.0 & 33.5 / 33.6 & 24.1 / 24.2 & 1.9 / 2.1 \\
        0.2 & 33.7 / 33.9 & 28.8 / 28.8 & 19.4 / 19.5 \\
        0.4 & 35.0 / 35.2 & 32.7 / 32.8 & 28.3 / 28.4 \\
        0.6 & 36.6 / 36.9 & 35.3 / 35.5 & 34.7 / 34.9 \\
        0.8 & {\bf 38.0} / {\bf 38.6} & 37.4 / 37.9 & 37.3 / 37.9 \\
        1.0 & 11.8 / 29.7 & 4.5 / 13.0 & non-convergence \\
        \bottomrule
    \end{tabular}
    \label{tab:assign}
\end{table}

\begin{table}[t]
    \centering
    \caption{The effect of various quality functions on COCO {\em val} set. `/' is used to distinguish between results without and with NMS. `Add' and `Mul' indicate two fusion functions.}
    \begin{tabular}{c|c|ccc}
        \toprule
        Method & $\alpha$ & mAP & AP$_{50}$ & AP$_{75}$ \\
        \midrule
        \multirow{3}{*}{Add} & 0.2 & 36.0 / 36.2 & {\bf 55.7} / 57.0 & 38.7 / 38.3 \\
         & 0.5 & 37.3 / 37.8 & 54.9 / 57.4 & 40.5 / 40.4 \\
         & 0.8 & 29.3 / 35.6 & 40.3 / 53.4 & 32.8 / 38.4 \\
        \midrule
        Mul & 0.8 & {\bf 38.0} / {\bf 38.6} & 55.2 / {\bf 57.6} & {\bf 41.4} / {\bf 41.3} \\
        \bottomrule
    \end{tabular}
    \label{tab:assign_form}
\end{table}

\noindent\textbf{Classification {\em vs.} regression.} 
The hyper-parameter $\alpha$, as shown in Eq.~\ref{eq:quality}, controls the ratio of the importance between classification and regression.
As reported in Tab.~\ref{tab:assign}, when $\alpha = 1$, the gap with NMS is not narrowed.
It could be attributed to the misalignment between the best positions for classification and regression.
When $\alpha = 0$, the assignment rule only relies on the predicted scores of classification.
Under this condition, the gap with NMS is considerably eliminated, but the absolute performance is still unsatisfactory, which could be caused by overfitting the sub-optimal initialization.
In contrast, with a proper fusion of classification and regression quality, the absolute performance is remarkably improved.

\begin{figure*}[t]
\centering
\begin{minipage}[t]{0.8\linewidth}
\centering
\includegraphics[width=\linewidth]{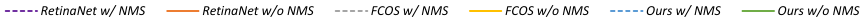}
\end{minipage} \\
\subfigure[mAP on COCO {\em val} set]{
\begin{minipage}[t]{0.33\linewidth}
\centering
\includegraphics[width=\linewidth]{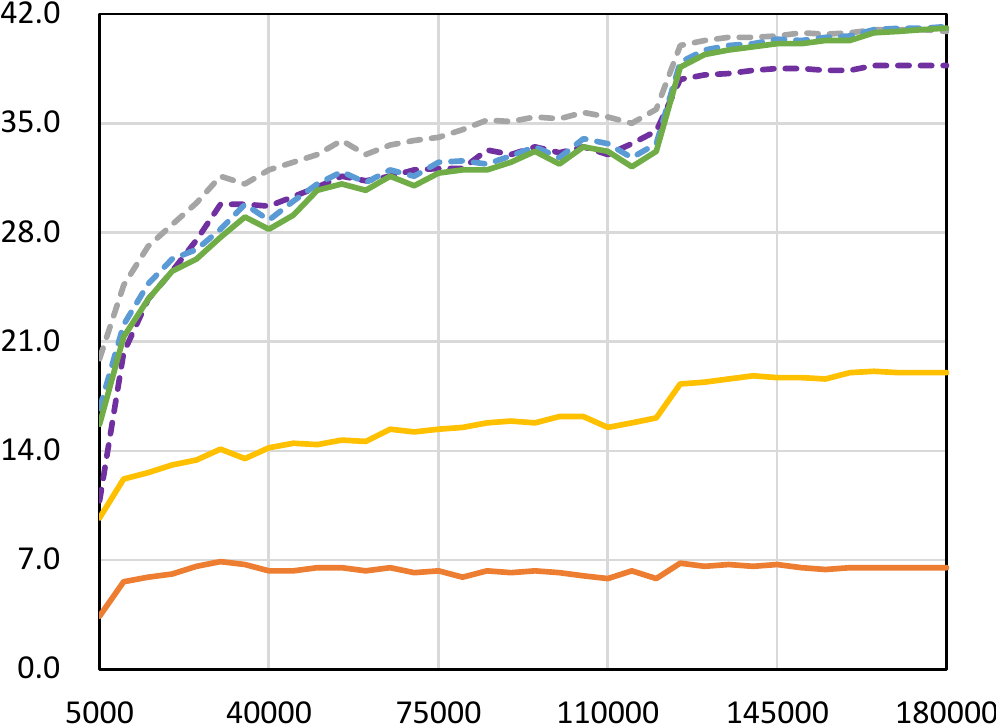}
\end{minipage}%
}%
\subfigure[mAR on COCO {\em val} set]{
\begin{minipage}[t]{0.33\linewidth}
\centering
\includegraphics[width=\linewidth]{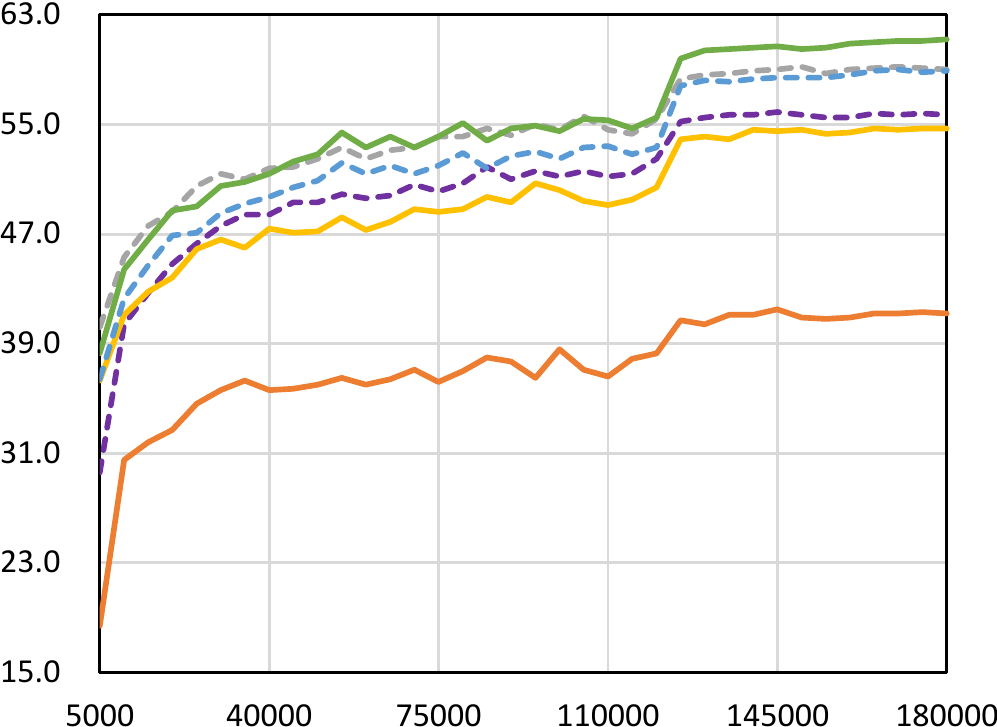}
\end{minipage}%
}
\subfigure[AP$_{50}$ on CrowdHuman {\em val} set]{
\begin{minipage}[t]{0.315\linewidth}
\centering
\includegraphics[width=\linewidth]{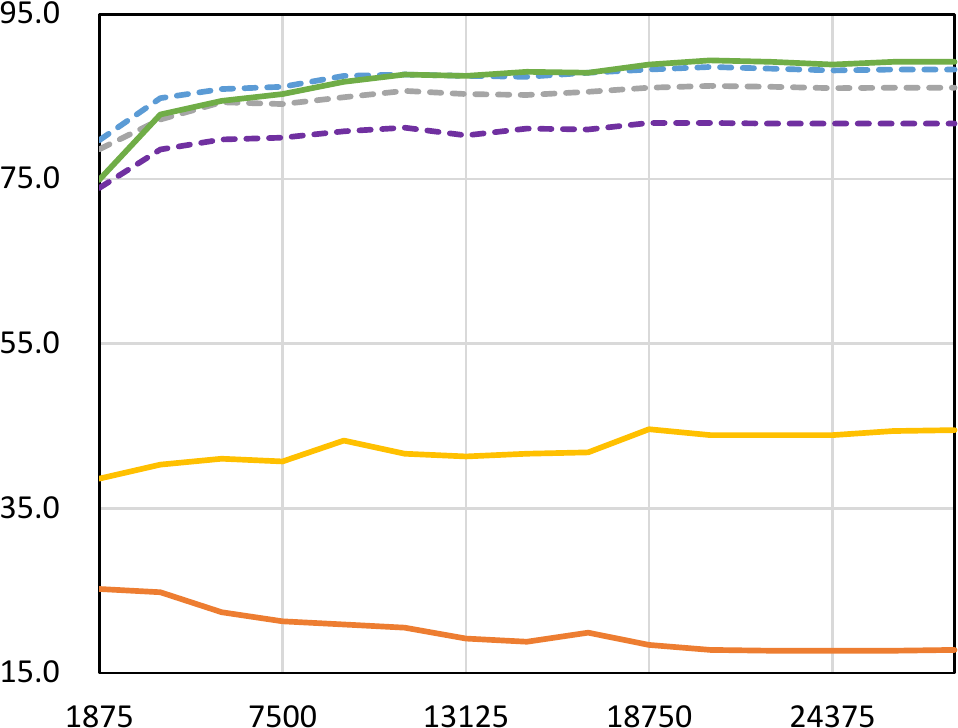}
\end{minipage}%
}
\caption{
The comparison graphs of performance {\em w.r.t.} training duration. The value of the horizontal axis corresponds to the training iterations. All the models are based on the ResNet-50 backbone. The threshold of NMS is set to 0.6.
}
\label{fig:map}
\end{figure*}

\noindent\textbf{Quality function.} 
We further explore the effect of different fusion methods on the quality function, {\em i.e.}, Eq.~\ref{eq:quality}.
As presented in Tab.~\ref{tab:assign_form}, the method called `Add' replaces the original quality function by $(1-\alpha) \cdot \hat{p}_{\pi(i)}(c_i) + \alpha \cdot \mathrm{IoU} \bigl( b_i, \hat{b}_{\pi(i)} \bigr)$, which has a similar form to \cite{li2020noisy}.
However, we find that the multiplication fusion, {\em i.e.}, `Mul',  is more suitable for the end-to-end detection, which achieves 0.7\% mAP absolute gains over the `Add' fusion method.

\begin{table}[t]
    \centering
    \caption{The effect of sub-modules in the proposed 3DMF module on COCO {\em val} set. `3DMF' and `Aux Loss' indicate using the 3D Max Filtering and the auxiliary loss, respectively. `/' is used to distinguish between results without and with NMS.}
\begin{threeparttable}
    \begin{tabular}{l|cc|c}
        \toprule
        Model & 3DMF & Aux Loss & mAP \\
        \midrule
        \multirow{3}{*}{FCOS~\cite{tian2019fcos}} & \xmark & \xmark & 19.0 / 40.9 \\  
         & \xmark & \cmark & 18.9 / {\bf 41.3} \\  
         & \cmark~\tnote{*} & \xmark & 38.7 / 40.0 \\  
        \midrule
        \multirow{3}{*}{Ours} & \xmark & \xmark & 38.0 / 38.6 \\
         & \cmark & \xmark & 39.8 / 40.0 \\
         & \cmark & \cmark & {\bf 41.1} / 41.2 \\
        \bottomrule
    \end{tabular}
    \begin{tablenotes}
        \small
        \item[*] We modify 3D Max Filtering as a post-processing.
    \end{tablenotes}
    \label{tab:improvement}
\end{threeparttable}
\end{table}
\begin{table}[t]
    \centering
    \caption{The effect of hyper-parameters in the proposed 3DMF module on COCO {\em val} set. $\tau = 0$ is equivalent to applying 2D Max Filtering to transform features on a single scale. `/' is used to distinguish between results without and with NMS.}
    \begin{tabular}{c|ccc}
        \toprule
         & $\phi=1$ & $\phi=3$ & $\phi=5$ \\  
        \midrule
        $\tau=0$ & 39.2 / 39.5 & 39.1 / 39.5 & 39.0 / 39.4 \\  
        $\tau=2$ & 39.0 / 39.3 & {\bf 39.8} / {\bf 40.0} & 39.3 / 39.5 \\  
        $\tau=4$ & 39.1 / 39.3 & 39.3 / 39.4 & 39.4 / 39.6 \\  
        \bottomrule
    \end{tabular}
    \label{tab:3d_max}
\end{table}

\subsubsection{3D Max Filtering}
\label{sec:3d_max_ablation}

\noindent\textbf{Components.} 
As shown in Tab.~\ref{tab:improvement}, without NMS post-processing, our end-to-end detector with POTO achieves 19.0\% mAP absolute gains over the vanilla FCOS.
By using the proposed 3DMF, the performance is further improved by 1.8\% mAP, and the gap with NMS is narrowed to 0.2\% mAP.
As shown in Fig.~\ref{fig:coco_vis}, the result shows the crucial role of the multi-scale and local-range suppression for end-to-end object detection.
The proposed auxiliary loss gives adequate supervision, making our detector obtain competitive performance against the FCOS with NMS.

\noindent\textbf{End-to-end.} 
To demonstrate the superiority of the end-to-end training manner, we replace the 2D Max Filtering of CenterNet~\cite{duan2019centernet} with the 3D Max Filtering as new post-processing for duplicate removal.
This post-processing is further adopted to the FCOS detector.
As shown in Tab.~\ref{tab:improvement}, the end-to-end manner achieves significant absolute gains by 1.1\% mAP.

\noindent\textbf{Kernel size.} 
As shown in Tab.~\ref{tab:3d_max}, we evaluate different settings of spatial range $\phi$ and scale range $\tau$ in the 3DMF.
When $\phi=3$ and $\tau=2$, our method obtains the highest performance on the COCO dataset.
This phenomenon reflects the duplicate predictions majorly come from a local region across adjacent scales, which is similar to the observation in Sec.~\ref{sec:3d_max}.

\noindent\textbf{Performance {\em w.r.t.} training duration.} 
As illustrated in Fig.~\ref{fig:map}(a), at the very beginning, the performance on COCO {\em val} set of our end-to-end detectors is inferior to the detectors with NMS.
As the training progressed, the performance gap becomes smaller and smaller.
After 180k training iterations, our method finally outperforms other detectors with NMS.
This phenomenon also occurs on CrowdHuman {\em val} set, which is shown in  Fig.~\ref{fig:map}(c).
Moreover, due to the removal of hand-designed post-processing, Fig.~\ref{fig:map}(b) demonstrates the superiority of our method in the recall rate against the NMS based methods.

\begin{table}[t]
    \centering
    \caption{The experiments of the proposed framework with larger backbone on COCO2017 {\em test-dev} set. The hyper-parameters of all the models follow the official settings.
    }
    \resizebox{\linewidth}{!}{
    \begin{tabular}{l|l|c|c}
        \toprule
        Backbone & Model & Epochs & mAP \\
        \midrule
        \multirow{5}{*}{ResNet-101} & RetinaNet~\cite{lin2017focal} & 36 & 41.0 \\
         & FCOS~\cite{tian2019fcos} & 36 & 43.1 \\
         & DETR~\cite{carion2020end} & 500 & 43.5 \\
        \cmidrule{2-4}
         & Ours (w/o NMS) & 36 & \bf43.6 \\
        \midrule
        \multirow{3}{*}{ResNeXt-101+DCN} & RetinaNet~\cite{lin2017focal} & 24 & 44.5 \\
         & FCOS~\cite{tian2019fcos} & 24 & 46.5 \\
        \cmidrule{2-4}
         & Ours (w/o NMS) & 24 & \bf47.6 \\
        \bottomrule
    \end{tabular}}
    \label{tab:sota}
\end{table}

\subsubsection{Larger Backbone}

To further demonstrate the robustness and effectiveness of our method, we provide experiments with larger backbones.
The detailed results are reported in Tab.~\ref{tab:sota}.
Concretely, when using the ResNet-101 as the backbone, our method is slightly superior to FCOS by 0.5\% mAP.
But when introducing more stronger backbone, {\em i.e.}, ResNeXt-101~\cite{xie2017aggregated} with deformable convolutions~\cite{zhu2019deformable}, our end-to-end detector achieves 1.1\% mAP absolute gains over the FCOS with NMS.
It might be attributed to the flexibly spatial modeling of deformable convolutions.
Moreover, the proposed 3DMF is efficient and easy to implement.
As shown in Tab.~\ref{tab:sota}, our 3DMF module only has a slightly computational overhead against the baseline detector with NMS.

\begin{table}[t]
    \centering
    \caption{The comparison of fully convolutional detectors on CrowdHuman {\em val} set. All models are based on the ResNet-50 backbone. `Aux' indicates the auxiliary loss.}
    \resizebox{\linewidth}{!}{
    \begin{tabular}{l|c|ccc}
        \toprule
        Method & Epochs & AP$_{50}$ & mMR & Recall \\
        \midrule
        RetinaNet~\cite{lin2017focal} & 32 & 81.7 & 57.6 & 88.6 \\
        FCOS~\cite{tian2019fcos} & 32 & 86.1 & 54.9 & 94.2 \\
        ATSS~\cite{zhang2020bridging} & 32 & 87.2 & 49.7 & 94.0 \\
        DETR~\cite{carion2020end} & 300 & 72.8 & 80.1 & 82.7 \\
        Ground-truth (w/ NMS) & - & - & - & 95.1 \\
        \midrule
        POTO & 32 & 88.5 & 52.2 & 96.3 \\
        POTO+3DMF & 32 & 88.8 & 51.0 & \bf 96.6 \\
        POTO+3DMF+Aux & 32 & \bf 89.1 & \bf 48.9 & 96.5 \\
        \bottomrule
    \end{tabular}}
    \label{tab:crowdhuman}
\end{table}

\subsection{Evaluation on CrowdHuman}

We evaluate our model on the CrowdHuman dataset~\cite{shao2018crowdhuman}, which is a large human detection dataset with various kinds of occlusions.
Compared with the COCO dataset, CrowdHuman has more complex and crowded scenes, giving severe challenges to conventional duplicate removal.
Our end-to-end detector is more robust and flexible in crowded scenes.
As shown in Tab.~\ref{tab:crowdhuman} and Fig.~\ref{fig:map}, our method significantly outperforms several state-of-the-art detectors with NMS, {\em e.g.}, 3.0\% mAP and 6.0\% mMR absolute gains over the FCOS.
Moreover, the recall rate of our method is even superior to the ground-truth boxes with NMS. 

\section{Conclusion}

This paper has presented a prediction-aware one-to-one label assignment and a 3D Max Filtering to bridge the gap between fully convolutional network and end-to-end object detection.
With the auxiliary loss, our end-to-end framework achieves superior performance against many state-of-the-art detectors with NMS on COCO and CrowdHuman datasets.
Our method is also demonstrated great potential in complex and crowded scenes, which may benefit many other instance-level tasks.

\section*{Acknowledgement}

This research was supported by National Key R\&D Program of China (No. 2017YFA0700800), National Natural Science Foundation of China (No. 61790563 and 61751401) and Beijing Academy of Artificial Intelligence (BAAI).

{\small
\bibliographystyle{ieee_fullname}
\bibliography{arxiv}
}

\clearpage

\appendix

\begin{figure*}[htbp]
\centering
\subfigure[Ground-truth]{
\begin{minipage}[t]{0.3\linewidth}
\centering
\includegraphics[width=\linewidth]{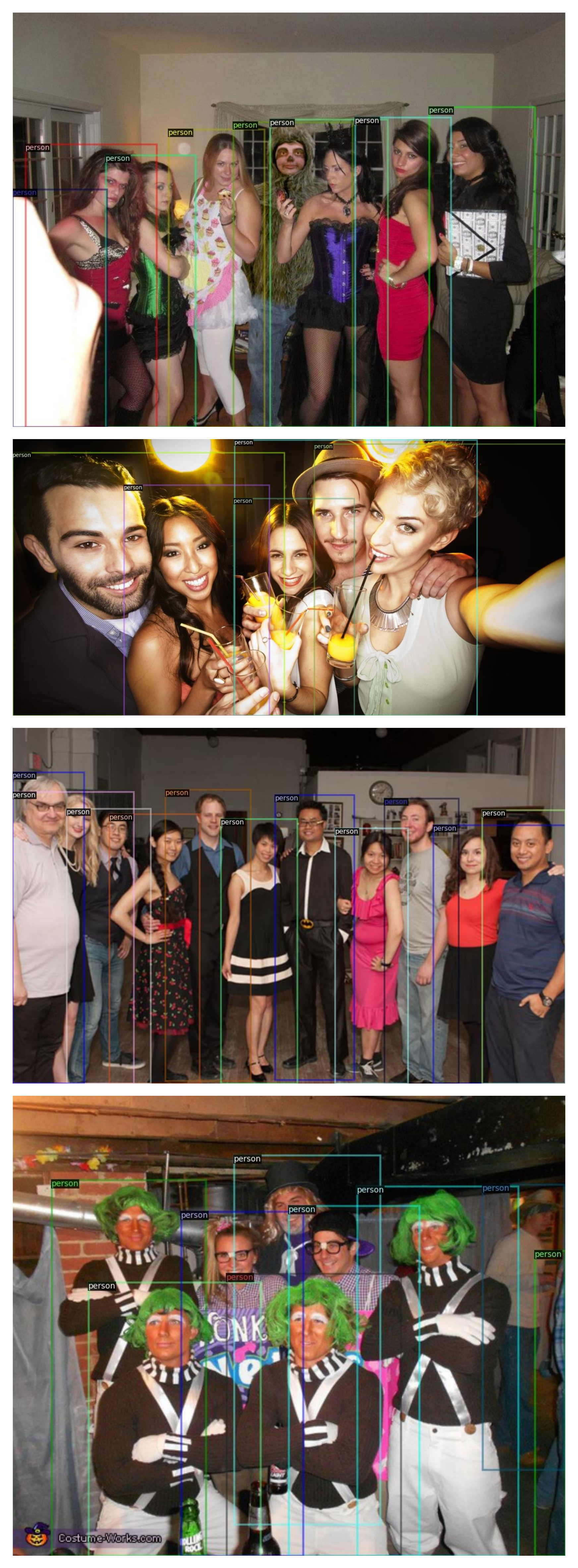}
\end{minipage}%
}%
\subfigure[FCOS baseline]{
\begin{minipage}[t]{0.3\linewidth}
\centering
\includegraphics[width=\linewidth]{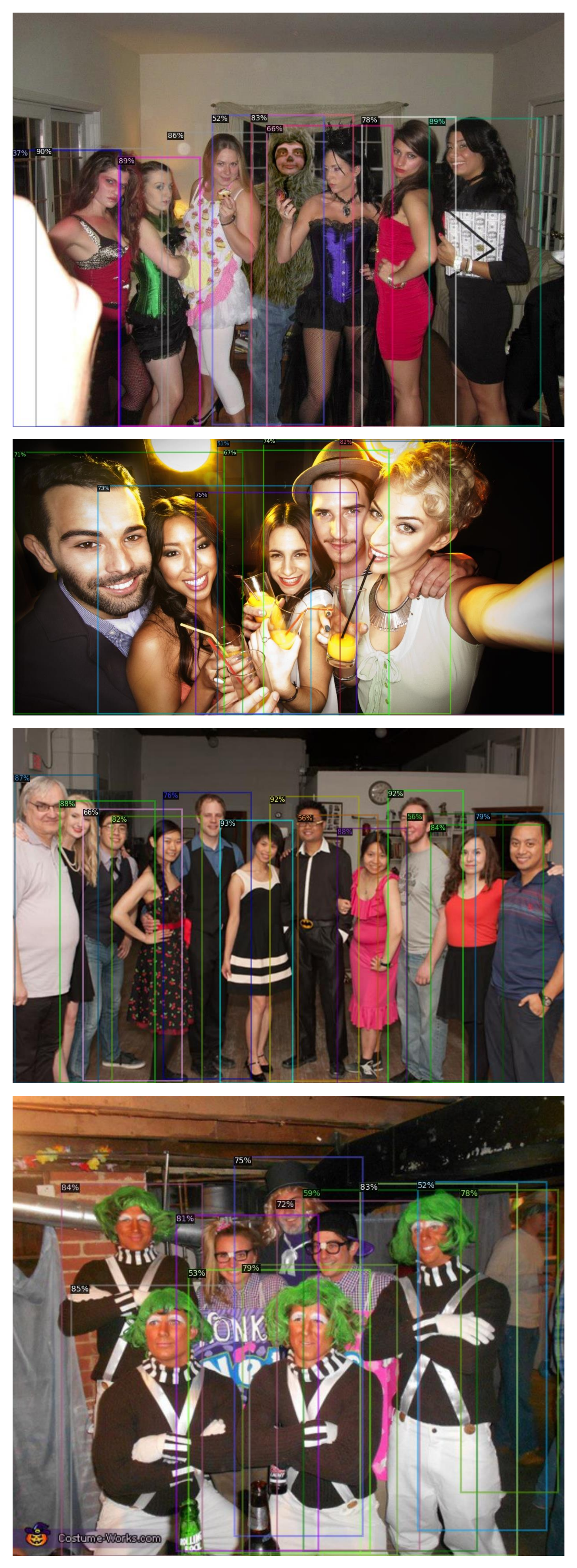} 
\end{minipage}%
}
\subfigure[Ours]{
\begin{minipage}[t]{0.3\linewidth}
\centering
\includegraphics[width=\linewidth]{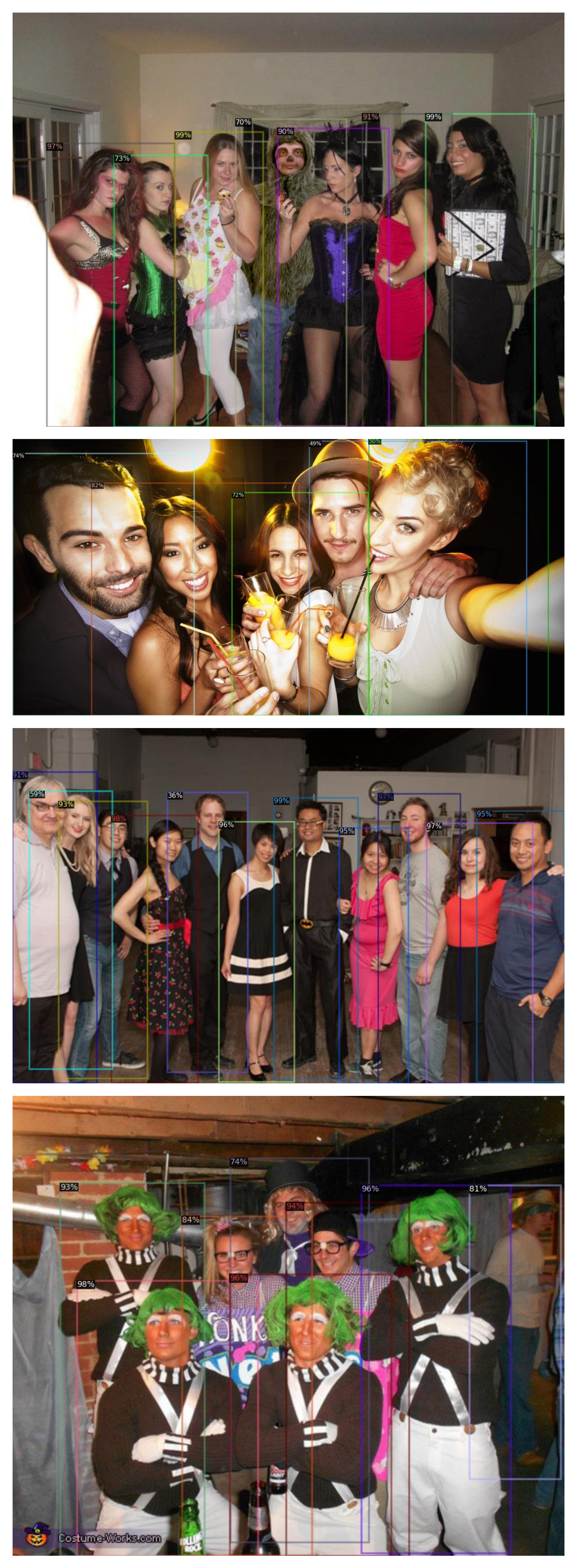}
\end{minipage}%
}
\caption{
The prediction visualizations of different detectors on CrowdHuman {\em val} set. Our method demonstrates superiority in the crowded scenes. All the models are based on the ResNet-50 backbone. The threshold of the classification score for visualization is set to 0.3.
}
\label{fig:crowdhuman}
\end{figure*}

\begin{figure*}[htbp]
\centering
\subfigure[Ground-truth]{
\begin{minipage}[t]{0.265\linewidth}
\centering
\includegraphics[width=\linewidth]{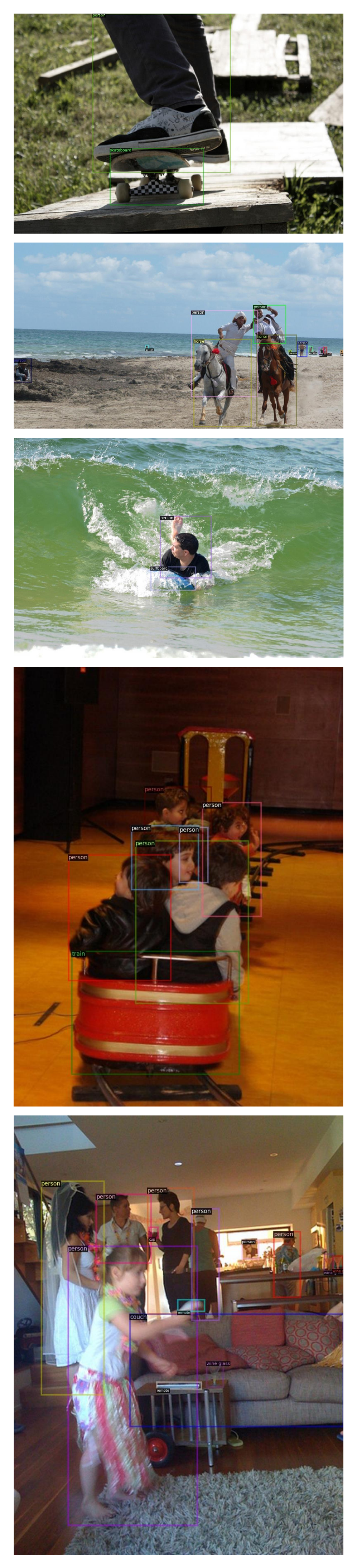}
\end{minipage}%
}
\subfigure[FCOS baseline]{
\begin{minipage}[t]{0.265\linewidth}
\centering
\includegraphics[width=\linewidth]{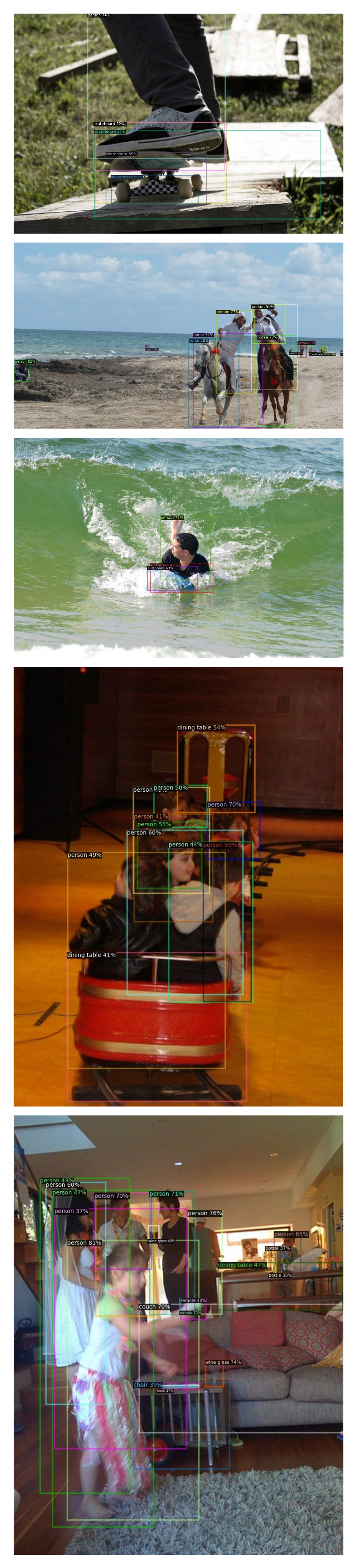} 
\end{minipage}%
}
\subfigure[Ours]{
\begin{minipage}[t]{0.265\linewidth}
\centering
\includegraphics[width=\linewidth]{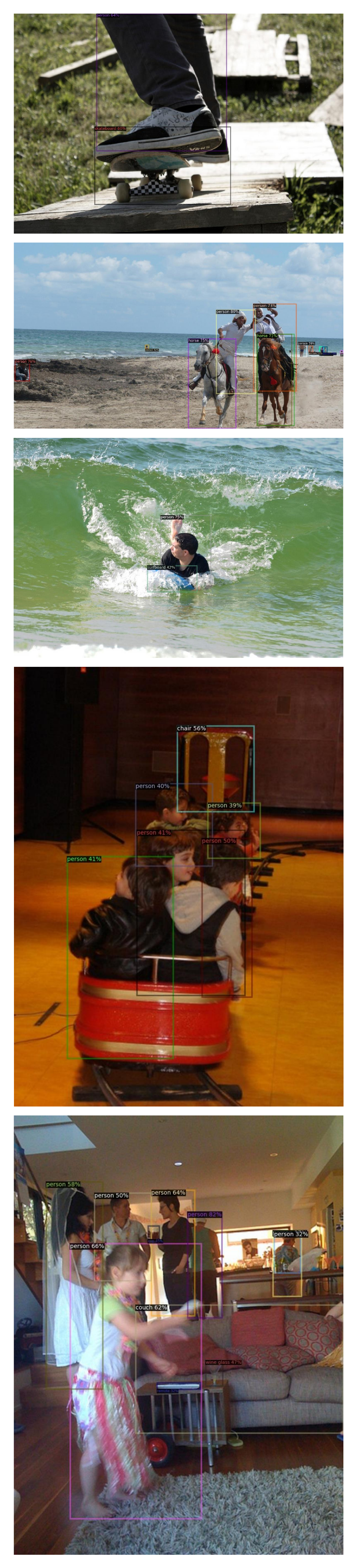}
\end{minipage}%
}
\caption{
The prediction visualizations of different detectors on COCO {\em val} set. Compared with the FCOS framework, our end-to-end detector obtains much fewer duplicate predictions, which is crucial for downstream instance-aware tasks. All the models are based on the ResNet-50 backbone. The threshold of the classification score for visualization is set to 0.3.
}
\label{fig:coco}
\end{figure*}

\clearpage

\section{Auxiliary Loss}

In this section, we evaluate different one-to-many label assignment rules for the auxiliary loss.
The detailed implementations are elaborated as follows:

\noindent\textbf{FCOS.}
We adopt the assignment rule in FCOS~\cite{tian2019fcos}.

\noindent\textbf{ATSS.}
We adopt the assignment rule in ATSS~\cite{zhang2020bridging}.

\noindent\textbf{Quality-ATSS.}
The rule is elaborated in Sec.~\ref{sec:aux_loss}.

\noindent\textbf{Quality-FCOS.}
Similar to FCOS, each ground-truth instance is assigned to the pixels in the pre-defined central area of a specific FPN stage. 
But the specific FPN stage is selected according to the proposed quality instead of the size of instances.

\noindent\textbf{Quality-Top-$k$.}
Each ground-truth instance is assigned to pixels with top-$k$ highest qualities over all the FPN stages. 
We set $k=9$ to align with other rules.

As shown in Tab.~\ref{tab:aux}, the results demonstrate the superiority of our proposed prediction-aware quality function over the hand-designed matching metrics.
Compared with the standard ATSS framework, the quality based rule can obtain 1.3\% mAP absolute gains.

\begin{table}[htbp]
    \centering
    \caption{The results of different one-to-many label assignment rules for the auxiliary loss on COCO {\em val} set. All the models are based on the ResNet-50 backbone. `/' is used to distinguish between results without and with NMS.}
    \begin{tabular}{l|ccc}
        \toprule
        Method & mAP & AP$_{50}$ & AP$_{75}$ \\
        \midrule
        None & 39.8 / 40.0 & 57.4 / 59.1 & 43.6 / 43.1 \\
        \midrule
        \multicolumn{4}{l}{\em Hand-designed} \\
        FCOS~\cite{tian2019fcos} & 39.4 / 39.8 & 57.0 / 59.1 & 43.4 / 43.0 \\
        ATSS~\cite{zhang2020bridging} & 39.8 / 40.1 & 57.5 / 59.5 & 44.1 / 43.4 \\
        \midrule
        \multicolumn{4}{l}{\em Prediction-aware} \\
        Quality-FCOS & 39.7 / 40.0 & 57.7 / 59.6 & 43.6 / 43.0 \\
        Quality-ATSS & \bf 41.1 / \bf 41.2 & \bf 59.0 / \bf 60.7 & \bf 45.4 / \bf 44.8 \\
        Quality-Top-$k$ & 40.7 / 41.0 & 58.7 / 60.4 & 44.9 / 44.3 \\
        \bottomrule
    \end{tabular}
    \label{tab:aux}
\end{table}

\section{Comparison to DETR}

As shown in Tab.~\ref{tab:detr_coco} and Tab.~\ref{tab:detr_crowdhuman}, we give the comparison of different methods based on ResNet-50 backbone, where the NMS is not utilized except for FCOS.

\begin{table}[htbp]
    \centering
    \caption{The comparison on COCO {\em val} set.}
    \resizebox{\linewidth}{!}{
    \begin{threeparttable}
    \begin{tabular}{l|c|c|ccc|c}
        \toprule
        Method & Epochs & $\rm{mAP}$ & $\rm{AP_s}$ & $\rm{AP_m}$ & $\rm{AP_l}$ & $\rm{\#Param}$\\
        \midrule
        DETR~\cite{carion2020end} & 500 & 42.0 & 20.5 & 45.8 & \bf 61.1 & 41.5 M \\
        FCOS~\cite{tian2019fcos} & 36 & 41.1 & 25.9 & 44.8 & 52.3 & 36.4 M \\
        Ours & 36 & 41.5 & \bf 26.4 & 44.7 & 52.8 & 37.0 M \\
        Ours\tnote{*} & 36 & \bf 43.5 & 26.3 & \bf 46.6 & 55.4 & 40.3 M \\  
        \bottomrule
    \end{tabular}
    \begin{tablenotes}
        \small
        \item[*] adopts two extra deformable convolutions in the head.
    \end{tablenotes}
    \end{threeparttable}}
    \label{tab:detr_coco}
\end{table}

\begin{table}[htbp]
    \centering
    \caption{The comparison on CrowdHuman {\em val} set.}
    \resizebox{\linewidth}{!}{\begin{threeparttable}
    \begin{tabular}{l|c|c|ccc}
        \toprule
        Method & Queries & Epochs & $\rm{AP_{50}}$ & $\rm{mMR}$ & $\rm{Recall}$ \\
        \midrule
        DETR~\cite{carion2020end} & 100 & 300 & 72.8 & 80.1 & 82.7 \\
        DETR & 200 & 300 & 78.8 & 66.3 & 90.2 \\
        DETR & 300 & 300 & 70.6 & 79.1 & 89.7 \\
        Ours & - & 32 & \bf 89.1 & \bf 48.9 & \bf 96.5 \\
        \bottomrule
    \end{tabular}
    \end{threeparttable}}
    \label{tab:detr_crowdhuman}
\end{table}

Compared with transformers, convolutions have been extensively tested in vision applications and have many variants for better performance than the DETR, {\em e.g.}, deformable convolutions~\cite{zhu2019deformable} in Tab.~\ref{tab:detr_coco}.
Moreover, as shown in Tab.~\ref{tab:detr_crowdhuman}, our framework has great advantages over the DETR~\cite{carion2020end} in convergence speed and crowded scenes.

\end{document}